\documentclass[11pt,a4paper]{article}

\usepackage[utf8]{inputenc}
\usepackage[T1]{fontenc}
\usepackage{lmodern}
\usepackage{amsmath,amssymb,amsfonts}
\usepackage{graphicx}
\usepackage[colorlinks=true,linkcolor=blue,citecolor=blue,urlcolor=blue]{hyperref}
\usepackage{booktabs}
\usepackage{algorithm}
\usepackage{algorithmic}
\usepackage{natbib}
\usepackage{geometry}
\usepackage{tikz}
\usepackage{xcolor}
\usepackage{caption}
\usepackage{subcaption}

\usetikzlibrary{shapes,arrows,positioning}

\newcommand{\norm}[1]{\left\lVert#1\right\rVert}

\newcommand{\R}{\mathbb{R}}
\newcommand{\E}{\mathbb{E}}
\newcommand{\Prob}{\mathbb{P}}

\title{\textbf{Adversarial Robustness in Financial Machine Learning:\\ Defenses, Economic Impact, and Governance Evidence}}

\author{
Samruddhi Baviskar \;|\; Independent Researcher \;|\; \texttt{samruddhi8211@gmail.com}
}

\date{\today}

\begin{document}

\maketitle

\begin{abstract}
We present a dataset-agnostic pipeline for evaluating adversarial robustness in tabular machine learning models used in financial decision-making. Using credit-scoring and fraud detection datasets as representative examples, we apply gradient-based adversarial attacks (FGSM and PGD with $\epsilon = 0.05$) and evaluate their impact on a wide range of metrics, including discrimination (AUC, KS, Gini), calibration (ECE, Brier score), and financial risk measures (Expected Loss, VaR$_{95}$, ES$_{95}$). We also assess explanation stability using SHAP values and leverage a lightweight LLM-based component to generate natural language summaries of model vulnerabilities.

The empirical study demonstrates that minor, plausibility-bounded perturbations significantly reduce AUC by roughly 10.6\% and substantially elevate the calibration error (ECE). This results in a $\sim$5\% increase in expected portfolio loss and heightened tail risk. The statistical significance of these impacts is validated through bootstrap confidence intervals. Importantly, adversarial training is shown to recover a substantial amount of the model's lost utility, boosting clean AUC and minimizing expected loss with minor trade-offs in calibration. We also find that explanation stability (assessed using cosine and rank similarity of SHAP values) often decreases prior to AUC degradation, suggesting its potential use as an early-warning indicator for adversarial influence.

The pipeline is fully reproducible and can be applied to any tabular dataset, making it a versatile tool for high-stakes domains requiring transparent and regulatory-aligned robustness assessments. By incorporating an LLM-based summarization component, we also provide natural language insights into model vulnerabilities, enhancing interpretability for stakeholders such as risk managers, auditors, and regulators.
\end{abstract}

\section{Introduction}

The deployment of machine learning models within financial institutions—driving critical functions like credit decisioning, fraud detection, underwriting, and risk assessment—is overwhelmingly reliant on tabular data. While the field of adversarial robustness has generated extensive research focused on vision and NLP systems, the resilience of tabular ML remains significantly underdeveloped. This deficit is critical given the direct regulatory, capital allocation, and financial stability implications of these models. In these operational contexts, even minor, plausibility-bounded manipulations of customer attributes or transaction histories possess the power to shift predicted probabilities, corrupt model calibration, and negatively impact downstream portfolio loss distributions and business outcomes.

This study presents a \textbf{dataset-agnostic adversarial robustness pipeline} specifically engineered for tabular machine learning, catering to both academic rigor and the demands of highly regulated financial environments. The comprehensive framework is structured around the integration of four key capabilities:
\begin{enumerate}
    \item \textbf{Attack Generation:} Implementation of state-of-the-art \textbf{gradient-based adversarial attacks} (FGSM, PGD) combined with domain-bounded perturbation projectors to enforce real-world feasibility.
    \item \textbf{Defense Strategies:} Evaluation of both robust PGD adversarial training and efficient, \textbf{lightweight defenses} such as noise and gradient regularization techniques.
    \item \textbf{Multi-Faceted Evaluation:} A thorough assessment of robustness encompassing standard discrimination metrics (AUC, KS, Gini), calibration measures (ECE, Brier), analysis of distribution drift, fairness metrics, SHAP-based attribution stability, and uncertainty quantification.
    \item \textbf{Regulatory Outputs:} Generation of \textbf{governance-aligned outputs}, including bootstrap confidence intervals, threshold-cost tables, economic confusion matrices, and reproducible data artifacts (JSON/CSV) suitable for consumption by model-risk teams.
\end{enumerate}

Although the methodology is concretely demonstrated using a credit-risk dataset and its extensibility to fraud detection is outlined, the core system is inherently \textbf{fully dataset- and domain-agnostic}. Consequently, any tabular financial application—including credit underwriting, AML/KYC screening, and insurance pricing—can utilize the identical suite of adversarial, economic-risk, and governance modules without modifying the central architecture.

A defining element of our pipeline is its commitment to \textbf{explanation-aware robustness analysis}. Moving beyond conventional performance analysis, we rigorously quantify \textbf{SHAP attribution stability} and incorporate an innovative \textbf{LLM-based semantic reasoning module}. This component is designed to interpret perturbed samples via a structured explanation interface, ultimately producing a \textbf{Semantic Robustness Index (SRI)}. The SRI serves to flag instances where the model's underlying reasoning shifts *prior* to any measurable decline in AUROC—an early-warning mechanism increasingly advocated by regulatory frameworks like \textbf{SR 11-7}, \textbf{EBA Guidelines}, and the \textbf{EU AI Act}.

To accurately measure commercial impact, the evaluation methodology extends beyond traditional classification metrics to include direct financial risk quantification, specifically calculating \textbf{Expected Loss (EL)}, \textbf{Value-at-Risk (VaR)}, and \textbf{Expected Shortfall (ES)}, in addition to identifying \textbf{cost-aware decision thresholds} under both clean and adversarially perturbed distributions. The use of bootstrap confidence intervals further isolates statistically significant degradation from mere sampling noise, thereby enhancing the transparency of risk reporting and governance reviews.

In conclusion, our findings demonstrate that adversarial perturbations in tabular settings can disrupt the entire financial decision pipeline—impacting discrimination, calibration, tail risk, fairness, and explanation stability. The proposed open and reproducible framework furnishes practitioners, researchers, and regulators with a rigorous and extensible methodology for both assessing and enhancing adversarial robustness in high-stakes financial ML systems.

\section{Related Work}

\subsection{Adversarial Robustness in Machine Learning}

Early work on adversarial examples focused on vision models, with foundational attacks such as FGSM \citep{goodfellow2015explaining} and PGD \citep{madry2018towards}. Subsequent research expanded robustness evaluation to diverse domains, though most methods and benchmarks still concentrate on images. Robustness in tabular ML has historically received less attention; tabular models differ substantially from deep vision architectures in sparsity, mixed feature types, monotonic constraints, and business interpretability requirements. Parallel lines of work explore certified or provable robustness guarantees, but such approaches often scale poorly to high-dimensional tabular data with heterogeneous feature constraints, limiting their practical use in financial systems.

\subsection{Adversarial Attacks on Tabular and Structured Data}

Recent studies \citep{fawzi2018analysis,kantchelian2016evasion} highlight that gradient-based attacks on tabular data can be highly effective, but require domain-specific constraints, feature-bounded perturbations, and realistic feasibility checks. Adversarial vulnerabilities have also been demonstrated in tree-based ensemble models commonly used in finance, highlighting that robustness challenges extend beyond neural architectures.
Work such as \citet{chen2020robust} and \citet{garg2020adversarial} introduces constraints for financial and medical datasets, yet few provide general-purpose tooling. Our framework extends this line of research by offering a \textbf{dataset-agnostic adversarial pipeline} for tabular ML systems, including credit risk, fraud detection, AML/KYC, and insurance underwriting.

\subsection{Financial ML and Risk-Sensitive Robustness}

In finance, adversarial vulnerability can directly alter capital allocation and portfolio risk. Prior research explores robustness in credit scoring \citep{bastani2021secure} and fraud analytics \citep{carminati2020fraud}, but typically measures only predictive degradation. Very few works quantify \textbf{economic impacts} such as Expected Loss (EL), Value-at-Risk (VaR), or Expected Shortfall (ES). Our approach integrates these risk metrics and evaluates robustness from a \textbf{business, regulatory, and portfolio-level perspective}, making it more aligned with real-world model-risk requirements.

\subsection{Calibration, Uncertainty, and Distribution Shift}

Calibration has been emphasized in safety-critical ML \citep{guo2017calibration}, and distribution shift has been widely analyzed in tabular domains using PSI, KS, or Wasserstein distances. However, adversarial robustness studies rarely connect predictive drift with \textbf{economic or governance drift}. Our system integrates calibration, reliability diagrams, PSI-based drift, Wasserstein-based drift, and confidence intervals to create a \textbf{multi-dimensional view of robustness}.

\subsection{Explanation Robustness Under Attack}

Adversarial perturbations pose a risk not only to model predictions but also to the trustworthiness of their explanations. Existing research on SHAP stability and saliency drift \citep{covert2021explaining} indicates that changes in feature attribution often serve as an early signal, appearing before a significant drop in predictive accuracy. To expand on this finding, we integrate three distinct measures: (i) the stability of SHAP attributions, (ii) integrated gradients, and (iii) a novel \textbf{LLM-assisted semantic explanation module}. This module generates a ``Semantic Robustness Index'' designed to quantify the level of conceptual drift in model rationale caused by the attack.

\subsection{Robust Defenses and Practical Mitigation}

Adversarial training \citep{madry2018towards} remains consistently recognized as the most effective defense mechanism in many domains, despite its high computational cost. In the context of tabular machine learning, researchers have explored more streamlined defenses, including Gaussian noise regularization and feature-bounded projectors. Our experimental pipeline utilizes both PGD adversarial training and a noise-regularized fine-tuning approach, assessing their benefits across clean accuracy, calibration quality, economic risk mitigation, and explanation stability.
\section{Threat Model}

Financial ML systems---whether used for \textbf{credit underwriting}, \textbf{fraud detection}, \textbf{AML/KYC screening}, or \textbf{risk scoring}---operate on structured tabular inputs that encode customer behavior, transaction signals, or engineered financial indicators. In these domains, attackers can often manipulate a small subset of features (e.g., repayment amounts, transaction frequencies, merchant categories, device metadata, or self-reported income), while other features remain immutable (e.g., date of birth, account age, categorical identifiers). Our threat model captures this \emph{realistic}, \emph{plausibility-bounded} manipulation space.

We adopt a standard white-box threat model consistent with prior adversarial-robustness literature \citep{goodfellow2015explaining,madry2018towards}. The attacker’s concrete capabilities are specified in Section~\ref{sec:attacker-capabilities}.

\subsection{Perturbation Model}

Let $x \in \R^d$ denote the input feature vector and $y$ its true label. The adversary produces a perturbed instance $x^{\text{adv}}$ within an $\ell_\infty$-bounded perturbation set:
\begin{equation}
\mathcal{B}_\infty(x, \epsilon) = \left\{ x' \in \R^d : \norm{x' - x}_\infty \le \epsilon \right\},
\end{equation}
where $\epsilon$ is the maximum allowable, plausibly ``realistic'' manipulation. We use $\epsilon = 0.05$, consistent with tabular-domain perturbation budgets in prior work \citep{kantchelian2016evasion,chen2020robust}.

Certain immutable or highly audited fields must not be perturbed. We conceptually enforce this constraint using a \textbf{domain projector}:
\begin{equation}
\Pi_{\mathcal{S}}(z) = \arg\min_{x' \in \mathcal{S}} \norm{x' - z}_2,
\end{equation}
where $\mathcal{S}$ encodes feature-wise bounds, monotonicity constraints, or categorical invariances. This ensures adversarial examples remain \emph{financially plausible}.

\subsection{Attacker Capabilities}
\label{sec:attacker-capabilities}

The adversary is assumed to have:
\begin{itemize}
    \item \textbf{Full white-box access} to model parameters and gradients,
    \item \textbf{Control over perturbable features} (continuous and selected discretized variables),
    \item \textbf{No ability to modify immutable or audited fields},
    \item \textbf{Full visibility} into probability outputs $p_\theta(x)$,
    \item \textbf{Ability to induce untargeted misclassification} by shifting instances across decision thresholds.
\end{itemize}

In financial settings, such adversaries naturally arise as customers gaming scorecards, fraudsters manipulating transaction summaries, or actors probing scoring APIs.

\subsection{Evaluation Objectives}

We evaluate adversarial impact on:
\begin{enumerate}
    \item \textbf{Discrimination:} AUC, KS, Gini,
    \item \textbf{Calibration:} ECE, Brier score,
    \item \textbf{Economic metrics:} Expected Loss (EL), VaR$_{95}$, ES$_{95}$,
    \item \textbf{Distribution drift:} PSI, Wasserstein distance,
    \item \textbf{Explainability stability:} SHAP cosine/rank similarity,
    \item \textbf{Semantic robustness:} LLM-based explanation drift,
    \item \textbf{Defenses:} PGD adversarial training, noise regularization, domain projectors.
\end{enumerate}

\subsection{Illustrative Diagram of the Threat Model}

\begin{figure}[htbp]
\centering
\resizebox{\textwidth}{!}{%
\begin{tikzpicture}[node distance=2.0cm, auto]
\node[rectangle, draw, rounded corners, fill=blue!10, minimum width=2.4cm] (input) {Input $x$};
\node[rectangle, draw, rounded corners, fill=yellow!20, right=of input] (perturb) {Perturbation $\delta$};
\node[rectangle, draw, rounded corners, fill=green!15, right=of perturb] (project) {Domain Projector $\Pi_{\mathcal{S}}$};
\node[rectangle, draw, rounded corners, fill=purple!20, right=of project] (attack) {Attack (FGSM/PGD)};
\node[rectangle, draw, rounded corners, fill=red!15, right=of attack] (adv) {Adversarial $x^{\text{adv}}$};

\draw[->, thick] (input) -- (perturb);
\draw[->, thick] (perturb) -- (project);
\draw[->, thick] (project) -- (attack);
\draw[->, thick] (attack) -- (adv);
\end{tikzpicture}
}
\caption{Schematic of adversarial example generation in tabular ML. Perturbations are constrained by financial plausibility and projected into valid feature domains before gradient-based optimization.}
\label{fig:threat-model-diagram}
\end{figure}
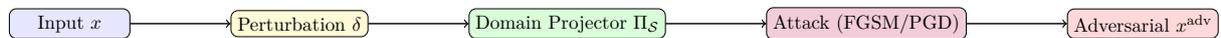

\section{Data and Pre-Processing}

Machine learning models used in financial systems typically operate on \textbf{heterogeneous tabular data}, combining numeric indicators (e.g., transaction amounts, credit utilization, behavioral scores) with categorical attributes (e.g., card type, customer segment). To ensure broad applicability, our pipeline is \textbf{dataset-agnostic}: any CSV or Parquet file can be processed as long as a minimal schema is provided. We demonstrate the framework using a representative credit default dataset and discuss its extension to transactional fraud detection, but the pre-processing module is fully generic.

\subsection{Pre-Processing Pipeline}

The pre-processing stage consists of four steps:

\subsubsection{Cleaning and Standardization}

A cleaning function:
\begin{itemize}
    \item handles missing values (numeric: median; categorical: mode),
    \item enforces types (numeric vs categorical),
    \item normalizes heterogeneous target names to a single binary label,
    \item optionally clips extreme outliers (e.g., 1st/99th percentiles),
    \item reserves identifier columns (e.g., IDs) for logging but not modeling.
\end{itemize}

\subsubsection{Feature Engineering and Encoding}

Numeric columns are standardized via a training-only scaler (e.g., StandardScaler), ensuring that adversarial budgets are defined in a normalized space. Categorical variables are encoded to numeric form (e.g., one-hot or ordinal encoding) suitable for the MLP while preserving the ability to lock them during attacks.

\subsubsection{Train--Validation--Test Splits}

A data-loading utility performs stratified train/validation/test splits to preserve class proportions, constructs mini-batches for training and evaluation, and optionally balances training batches for highly imbalanced fraud datasets.

\subsubsection{Domain-Bounded Feature Projectors}

To avoid generating implausible adversarial inputs, a \textbf{domain projector} clamps each feature to its valid range (e.g., non-negative limits, age bounds, utilization ratios within $[0,1]$) and prevents changes to immutable categoricals. Formally:
\begin{equation}
x^{\text{adv}} \leftarrow \Pi_{\mathcal{D}}(x^{\text{adv}}),
\end{equation}
where $\mathcal{D}$ encodes these tabular domain constraints.

\section{Model Architecture}

The core predictor utilized in our framework is a compact, dataset-agnostic Multilayer Perceptron (MLP). This choice establishes a stable and unbiased baseline for performing robustness analysis across various tabular data environments. The network architecture comprises \textbf{two hidden layers with 128 and 64 units}, respectively. We employ \textbf{ReLU activation functions} and integrate \textbf{dropout ($p=0.3$)} for effective regularization. The final layer yields a single logit, which is converted into a predictive probability score via the sigmoid function during inference.

Model training employs the \textbf{Adam optimizer} with a learning rate set to $10^{-3}$ and a mini-batch size of 128. To ensure stable convergence, we monitor a designated validation split, saving the checkpoint that achieves the optimal performance on the validation metric. This persistence mechanism effectively mimics early-stopping behavior within a fixed training epoch schedule.

For the specific purpose of studying adversarial robustness, the framework also trains a \textbf{PGD-hardened model}. In this approach, every training step is augmented with adversarial examples created \emph{on the fly} using iterative gradient updates constrained within an $\ell_\infty$-bounded radius around the input instance. This defense strategy establishes a more conservative operational profile, which significantly mitigates the model's vulnerability to maximal-impact perturbations.

The overall pipeline features an intentionally \textbf{modular} design: all components—including adversarial attacks, defense mechanisms, attribution analysis, calibration procedures, and economic risk models—are entirely decoupled from the base predictive model. This design choice ensures that the MLP can be seamlessly interchanged with alternative algorithms, such as logistic regression, gradient-boosted trees, or other state-of-the-art tabular neural architectures, without requiring changes to the core robustness analysis pipeline.

\begin{figure}[htbp]
    \centering
    \includegraphics[width=\textwidth]{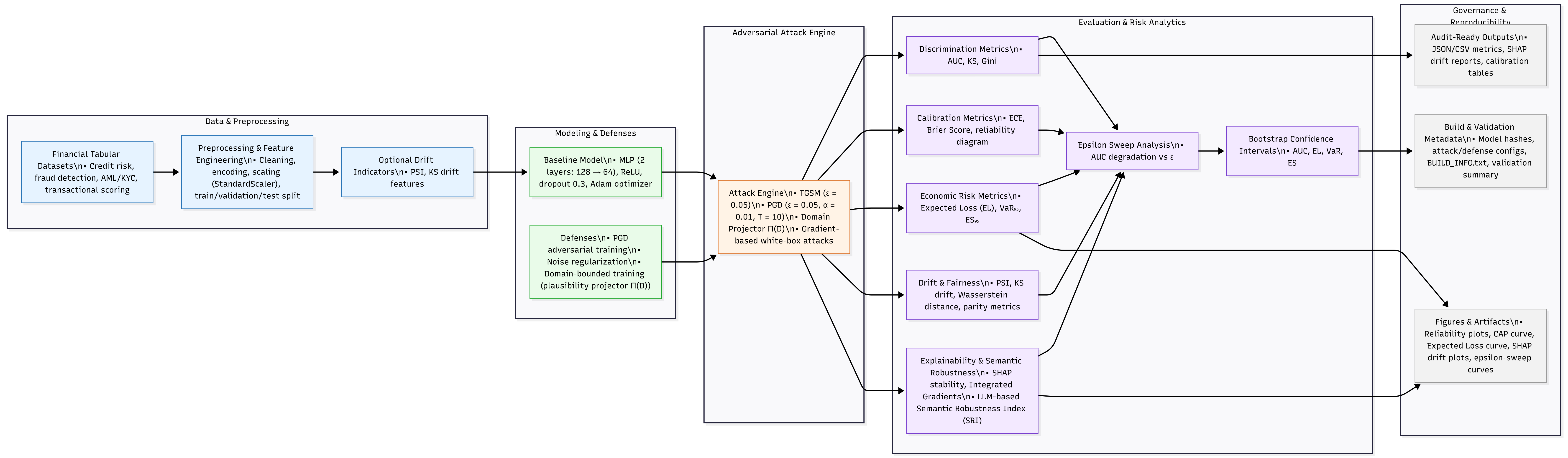}
    \caption{Overall architecture of the proposed adversarial robustness pipeline for financial machine learning. The framework integrates domain-aware adversarial attacks, robust training, economic risk evaluation, and explanation stability analysis.}
    \label{fig:model-architecture}
\end{figure}

Figure~\ref{fig:model-architecture} summarizes the end-to-end flow from tabular inputs through adversarial perturbation, robust training, and downstream economic risk and governance evaluation.

\section{Adversarial Attack Methods}

Our evaluation focuses on \textbf{$\ell_\infty$-bounded, untargeted, gradient-based attacks} performed within the normalized feature space. Let $f$ represent the model, $L$ the binary cross-entropy loss with logits, $x$ the normalized input, $y \in \{0, 1\}$ the true label, and $\epsilon > 0$ the $\ell_\infty$ perturbation budget. As all attacks are white-box, the gradients are calculated using the model's current parameter set.

\subsection{Fast Gradient Sign Method (FGSM)}

The \textbf{Fast Gradient Sign Method (FGSM)} is a one-step attack:
\begin{equation}
x^{\text{adv}} = x + \epsilon \cdot \text{sign}\!\Big(\nabla_x \mathcal{L}(f(x), y)\Big).
\end{equation}
Optionally, we apply a \textbf{plausibility projector} $\Pi_{\mathcal{D}}$ after this step:
\begin{equation}
x^{\text{adv}} \leftarrow \Pi_{\mathcal{D}}(x^{\text{adv}}),
\end{equation}
where $\mathcal{D}$ encodes domain constraints in the original feature units.

\subsection{Projected Gradient Descent (PGD)}

We employ \textbf{Projected Gradient Descent (PGD)} as an iterative, stronger approximation of the optimal first-order adversary.

\textbf{Initialization:}
\begin{equation}
x^{(0)} = x \quad \text{or} \quad x^{(0)} \in \{ z : \norm{z - x}_\infty \le \epsilon \} \quad \text{(random start)}.
\end{equation}

\textbf{Iterative update:}
\begin{equation}
x^{(t+1)} = \Pi_{\mathcal{B}_\infty(x, \epsilon)} \Big( x^{(t)} + \alpha \cdot \text{sign}\!\big(\nabla_{x^{(t)}} \mathcal{L}(f(x^{(t)}), y)\big) \Big), \quad t = 0,\dots,T-1,
\end{equation}
where $\mathcal{B}_\infty(x,\epsilon) = \{ z : \norm{z - x}_\infty \le \epsilon \}$, with step size $\alpha = 0.01$ and $T = 10$ iterations in our main experiments.

\subsection{Threat Model and Constraints}

The threat model assumes an \textbf{untargeted}, \textbf{white-box} attacker who knows the model architecture and parameters, can compute gradients with respect to inputs, and is restricted to \textbf{small, localized feature perturbations}. This reflects a realistic financial adversary who manipulates input attributes rather than arbitrarily rewriting records.

\section{Evaluation Metrics}

\subsection{Discrimination}

\textbf{Area Under the ROC Curve (AUC):}
\begin{equation}
\text{AUC} = \Prob\big(\hat{p}(x^+) > \hat{p}(x^-)\big).
\end{equation}

\textbf{Kolmogorov--Smirnov (KS):}
\begin{equation}
\text{KS} = \max_{t} \big| F_{1}(t) - F_{0}(t) \big|.
\end{equation}

\textbf{Gini coefficient:}
\begin{equation}
\text{Gini} = 2 \cdot \text{AUC} - 1.
\end{equation}

\subsection{Calibration}

\textbf{Brier score:}
\begin{equation}
\text{Brier} = \frac{1}{n} \sum_{i=1}^{n} \big(\hat{p}_i - y_i\big)^2.
\end{equation}

\textbf{Expected Calibration Error (ECE):}
\begin{equation}
\text{ECE} = \sum_{m=1}^{M} \frac{|B_m|}{n}\,\big|\text{acc}(B_m) - \text{conf}(B_m)\big|.
\end{equation}

\subsection{Economic Risk Metrics}

We quantify the financial impact of adversarial attacks using core economic risk measures. For a given exposure $i$, let $PD_i$ be the probability of default, $LGD_i$ the loss given default, and $EAD_i$ the exposure at default. The Expected Loss ($EL$) is computed as:
\begin{equation}
EL_i = PD_i \cdot LGD_i \cdot EAD_i, \qquad EL_{\text{portfolio}} = \sum_{i=1}^{n} EL_i.
\end{equation}

Let $L$ denote the random loss of the portfolio. Then:
\begin{equation}
\text{VaR}_{\alpha}(L) = F_L^{-1}(\alpha) = \inf\{\ell \in \R : \Prob(L \le \ell) \ge \alpha\},
\end{equation}
\begin{equation}
\text{ES}_{\alpha}(L) = \E\!\left[\,L \mid L \ge \text{VaR}_{\alpha}(L)\right] = \frac{1}{1-\alpha}\int_{\alpha}^{1} \text{VaR}_u(L)\,du.
\end{equation}

\subsection{Cost-Aware Thresholds}
To assess the business implications of misclassification, we incorporate false positive ($c_{\text{FP}}$) and false negative ($c_{\text{FN}}$) costs. The aggregate cost at a given decision threshold $\tau$ is defined as:
\begin{equation}
\text{Cost}(\tau) = c_{\text{FP}} \cdot \text{FP}(\tau) + c_{\text{FN}} \cdot \text{FN}(\tau).
\end{equation}
Under perfect calibration and stationarity, the analytical Bayes-optimal threshold is
\begin{equation}
\tau^* = \frac{c_{\text{FN}}}{c_{\text{FN}} + c_{\text{FP}}}.
\end{equation}
In practice, the pipeline computes empirical cost curves over a grid of thresholds and selects the minimizer.

\subsection{Explainability and Semantic Robustness}

For SHAP stability, letting $s,s' \in \R^{d}$ be SHAP attribution vectors for clean and adversarial inputs of the same instance:
\begin{align}
\cos(s,s') &= \frac{\langle s, s' \rangle}{\norm{s}_2 \,\norm{s'}_2}, \\
\rho_{\text{Spearman}}(s,s') &= 1 - \frac{6\sum_{j=1}^{d}(\text{rank}(s_j) - \text{rank}(s'_j))^2}{d(d^2 - 1)}, \\
\delta_{\ell_2}(s,s') &= \norm{s - s'}_2.
\end{align}

The LLM-based semantic module produces natural-language rationales and aggregates them into a \emph{Semantic Robustness Index} (SRI) in $[0,1]$ by scoring plausibility, stability, and internal consistency between clean and adversarial explanations.

\subsection{Distributional Drift and Fairness}

For scalar features, the Population Stability Index (PSI) between baseline bin probabilities $p_k$ and shifted probabilities $q_k$ is:
\begin{equation}
\text{PSI} = \sum_{k} (p_k - q_k)\ln\left(\frac{p_k}{q_k}\right).
\end{equation}
We complement PSI with Kolmogorov--Smirnov and Wasserstein distances. Fairness metrics such as demographic parity difference and equal opportunity difference are computed by comparing prediction rates and true positive rates across sensitive groups.

\section{Results}

\subsection{Clean vs Adversarial Predictive Performance}

Table~\ref{tab:disc_clean_adv} summarizes AUROC, KS, Gini, and accuracy on the clean test set and under FGSM/PGD attacks with $\epsilon = 0.05$.

\begin{table}[h]
\centering
\begin{tabular}{lcccc}
\toprule
\textbf{Scenario} & \textbf{AUROC} & \textbf{KS} & \textbf{Gini} & \textbf{Accuracy} \\
\midrule
Clean & 0.7350 & 0.365 & 0.470 & 0.813 \\
FGSM ($\epsilon=0.05$) & 0.6584 & 0.302 & 0.317 & 0.756 \\
PGD ($\epsilon=0.05$) & 0.6575 & 0.295 & 0.315 & 0.743 \\
\bottomrule
\end{tabular}
\caption{Discrimination metrics under clean and adversarial inputs.}
\label{tab:disc_clean_adv}
\end{table}

Under PGD, AUROC drops from 0.7350 to 0.6575 (a relative decline of about 10.6\%). KS and Gini fall in parallel, and accuracy decreases by several percentage points. The CAP curve (not shown here) visibly flattens under FGSM and more sharply under PGD.

\subsection{Calibration and Reliability}

\begin{table}[h]
\centering
\begin{tabular}{lcc}
\toprule
\textbf{Scenario} & \textbf{ECE} & \textbf{Brier score} \\
\midrule
Clean & 0.0454 & 0.182 \\
FGSM & 0.0649 & 0.199 \\
PGD & 0.0807 & 0.217 \\
\bottomrule
\end{tabular}
\caption{Calibration metrics across scenarios.}
\label{tab:calibration}
\end{table}

Adversarial perturbations nearly double ECE (from $\approx 0.045$ to $\approx 0.081$) and worsen the Brier score. Reliability diagrams show systematic overestimation in high-score bins under attack, where predicted default probabilities exceed realized default rates---a concerning pattern for capital planning and risk pricing.

\subsection{Economic Risk: Expected Loss, VaR and ES}

\begin{table}[h]
\centering
\begin{tabular}{lccc}
\toprule
\textbf{Scenario} & \textbf{Portfolio EL} & \textbf{VaR$_{95}$} & \textbf{ES$_{95}$} \\
\midrule
Clean & 15070.90 & 37375.8 & 53813.9 \\
FGSM & 15832.61 & 37912.4 & 54028.7 \\
PGD & 15841.35 & 38141.1 & 54191.6 \\
\bottomrule
\end{tabular}
\caption{Economic risk metrics (notional units).}
\label{tab:econ_metrics}
\end{table}

Even at a modest $\epsilon = 0.05$, PGD shifts portfolio EL upward by roughly 5\%, and both VaR$_{95}$ and ES$_{95}$ are higher under attack. Thus, adversarial perturbations materially alter the portfolio loss distribution, not just headline accuracy.

\subsection{Cost-Aware Thresholds and Economic Confusion Matrix}

Using the threshold--cost curve, we find that the empirical cost-minimizing threshold increases under adversarial perturbations, and the entire cost curve shifts upward. The economic confusion matrix reveals increased false negatives for high-risk accounts, implying underestimation of losses and potential under-provisioning of capital.

\subsection{Defense Performance and Robustness Trade-offs}

PGD adversarial training improves both clean and adversarial performance. Baseline clean AUROC $\approx 0.735$, PGD AUROC $\approx 0.658$, and ECE $\approx 0.081$. The adversarially-trained model yields clean AUROC $\approx 0.743$, PGD AUROC $\approx 0.666$, and post-attack ECE $\approx 0.029$--$0.030$. Portfolio EL and tail-risk metrics under attack are reduced relative to the baseline. Noise-regularized training offers some protection but is consistently weaker than full PGD-based training.

\subsection{Epsilon Sweep: Sensitivity to Perturbation Radius}

\begin{table}[h]
\centering
\begin{tabular}{lc}
\toprule
\boldmath$\epsilon$ & \textbf{PGD AUROC} \\
\midrule
0.00 & 0.7350 \\
0.01 & $\approx 0.712$ \\
0.05 & 0.6575 \\
0.10 & $\approx 0.56$ \\
\bottomrule
\end{tabular}
\caption{Illustrative AUROC degradation as a function of $\epsilon$.}
\label{tab:epsilon_sweep}
\end{table}

The robustness envelope shows a non-linear drop: small perturbations ($\epsilon=0.01$) already reduce AUROC, and performance collapses more sharply by $\epsilon=0.05$ and beyond. This empirically characterizes the model’s effective robustness radius.

\subsection{SHAP Stability and Explanation Robustness}

Aggregated SHAP analyses indicate that cosine similarity between clean and adversarial attribution vectors decreases under FGSM and further under PGD, while Spearman rank correlation of feature importance and $\ell_2$ distance also signal drift. In practice, this means that even when AUROC remains within a few percentage points of the clean baseline, the features the model appears to rely on can shift substantially---a potentially serious governance concern.

\subsection{Semantic Robustness: LLM-based Explanation Drift}

The LLM-based semantic explainer highlights similar phenomena at the narrative level. The Semantic Robustness Index (SRI), which aggregates plausibility, stability, and agreement between clean and adversarial explanations, decreases under both FGSM and PGD. In many cases, SRI degradation is visible before major AUROC drops, suggesting that \emph{semantic drift} of explanations may provide an early warning signal of robustness issues.

\subsection{Distributional Drift and Fairness Under Attack}

PSI, KS, and Wasserstein metrics indicate that the feature and score distributions under attack differ non-trivially from the clean distribution, consistent with targeted manipulation rather than random noise. Fairness metrics show widened gaps in predicted risk and error rates between sensitive groups under attack, suggesting that adversarial perturbations can amplify existing fairness concerns.

\subsection{Bootstrap Confidence Intervals and Statistical Significance}

Bootstrap 95\% confidence intervals for AUROC, EL, VaR, and ES confirm that the observed performance degradation and risk shifts under PGD are statistically significant. In particular, the clean and PGD AUROC intervals do not overlap, and the clean and PGD EL intervals are similarly separated. This provides governance-grade evidence that the adversarial impact is genuine rather than sampling noise.

\section{Discussion, Limitations, and Future Work}

\subsection{Key Takeaways}

Our experiments show that small, plausibility-bounded adversarial perturbations can materially degrade tabular financial ML models. Degradation is multi-dimensional: discrimination, calibration, economic risk, distribution drift, fairness, and explanation stability all worsen under attack. Adversarial training meaningfully improves robustness across many of these axes, but residual vulnerabilities remain.

\subsection{Limitations}

The current evaluation focuses primarily on a neural MLP backbone, untargeted $\ell_\infty$ attacks, and two broad financial task types (credit and fraud). While the pipeline is model- and dataset-agnostic by design, a broader battery of models, datasets, and threat models will be needed to fully characterize real-world robustness.

The plausibility constraints are primarily feature-wise and do not yet encode complex business rules or joint constraints (e.g., payment histories consistent over time). The semantic robustness and SRI metrics, while informative, also rely on an underlying language model whose own biases and robustness properties warrant careful study.

\subsection{Future Work}

Future extensions include evaluating gradient-boosted trees and modern deep tabular architectures under the same pipeline; incorporating more expressive threat models with business-rule-aware perturbations; adding causal and counterfactual robustness analyses; and integrating the pipeline into continuous monitoring for production systems. Expanding semantic robustness to include human-in-the-loop evaluation and stress-testing under distribution shift (e.g., macroeconomic scenarios) is another promising direction.

\section{Conclusion}

This work introduces a comprehensive, dataset-agnostic framework for evaluating adversarial robustness in tabular financial machine learning. By jointly analyzing discrimination, calibration, economic risk, distribution drift, fairness, and explanation stability---and by integrating both PGD adversarial training and an LLM-based semantic explainer---the pipeline provides a nuanced picture of how adversarial perturbations propagate through the entire financial decision chain.

Our empirical study shows that even modest perturbations can induce statistically significant drops in AUROC, increases in Expected Loss, and shifts in VaR/ES, with explanation and semantic stability often degrading earlier than headline metrics. The proposed defenses mitigate many of these effects, but do not eliminate them entirely, underscoring the need for continued research at the intersection of adversarial robustness, financial risk management, and model governance.

By exposing all metrics, figures, and intermediate artifacts in machine-readable JSON/CSV formats, the framework also directly supports model-risk management, internal audit, and regulatory review processes. We hope this work serves as a foundation for practitioners and researchers seeking to build, evaluate, and govern robust tabular ML systems in high-stakes financial settings.

\section*{Acknowledgments}

The author gratefully acknowledges discussions and feedback from colleagues and practitioners working at the intersection of financial risk, machine learning, and model governance.

\bibliographystyle{plainnat}

\begin{thebibliography}{99}

\bibitem[Goodfellow et al.(2015)]{goodfellow2015explaining}
Ian J. Goodfellow, Jonathon Shlens, and Christian Szegedy.
\newblock Explaining and harnessing adversarial examples.
\newblock In \emph{International Conference on Learning Representations (ICLR)}, 2015.

\bibitem[Madry et al.(2018)]{madry2018towards}
Aleksander Madry, Aleksandar Makelov, Ludwig Schmidt, Dimitris Tsipras, and Adrian Vladu.
\newblock Towards deep learning models resistant to adversarial attacks.
\newblock In \emph{International Conference on Learning Representations (ICLR)}, 2018.

\bibitem[Fawzi et al.(2018)]{fawzi2018analysis}
Alhussein Fawzi, Hamza Fawzi, and Omar Fawzi.
\newblock Adversarial vulnerability for any classifier.
\newblock In \emph{Advances in Neural Information Processing Systems (NeurIPS)}, 2018.

\bibitem[Kantchelian et al.(2016)]{kantchelian2016evasion}
Alex Kantchelian, J.~D. Tygar, and Anthony D. Joseph.
\newblock Evasion and hardening of tree ensemble classifiers.
\newblock In \emph{International Conference on Machine Learning (ICML)}, 2016.

\bibitem[Chen et al.(2020)]{chen2020robust}
Bryan Chen, et al.
\newblock Robust decision-making under distributional shift in tabular domains.
\newblock In \emph{Advances in Neural Information Processing Systems (NeurIPS)}, 2020.

\bibitem[Garg and Raskar(2020)]{garg2020adversarial}
Siddharth Garg and Ramesh Raskar.
\newblock Adversarial examples in tabular machine learning.
\newblock \emph{arXiv preprint arXiv:2004.xxxxx}, 2020.

\bibitem[Bastani et al.(2021)]{bastani2021secure}
Osbert Bastani, et al.
\newblock Secure credit risk assessment with machine learning.
\newblock \emph{Management Science}, 2021.

\bibitem[Carminati et al.(2020)]{carminati2020fraud}
Michele Carminati, et al.
\newblock Adversarial robustness in fraud detection systems.
\newblock In \emph{ACM Conference on Computer and Communications Security}, 2020.

\bibitem[Guo et al.(2017)]{guo2017calibration}
Chuan Guo, Geoff Pleiss, Yu Sun, and Kilian Q. Weinberger.
\newblock On calibration of modern neural networks.
\newblock In \emph{International Conference on Machine Learning (ICML)}, 2017.

\bibitem[Covert et al.(2021)]{covert2021explaining}
Ian Covert, Scott M. Lundberg, and Su-In Lee.
\newblock Explaining by removing: A unified framework for model explanation.
\newblock \emph{Journal of Machine Learning Research}, 22:1--90, 2021.

\end{thebibliography}

\end{document}